\documentclass[10pt,twocolumn,letterpaper]{article}

\usepackage[pagenumbers]{cvpr} % To force page numbers, e.g. for an arXiv version

\usepackage{booktabs, multirow} % for borders and merged ranges
\usepackage{soul}% for underlines
\usepackage[dvipsnames,svgnames]{xcolor} % for extended list of colors
\usepackage{changepage,threeparttable} % for wide tables
\usepackage{graphicx}
\usepackage{subcaption}
\usepackage{dblfloatfix}
\usepackage{makecell}

\newif\ifshowcomments
\showcommentsfalse
\newcommand{\vale}[1]{\ifshowcomments\textbf{\textcolor{Mulberry}{[V: #1]}}\fi}

\newif\ifshowappendix
\showappendixtrue

\definecolor{cvprblue}{rgb}{0.21,0.49,0.74}
\usepackage[pagebackref,breaklinks,colorlinks,allcolors=cvprblue]{hyperref}

\def\paperID{*****} % *** Enter the Paper ID here
\def\confName{CVPR}
\def\confYear{2026}

\title{Making the Discrete Continuous: Synthetic RAW Augmentations for Fine-Grained Evaluation of Person Detection Performance in Low Light}

\author{Valeria Pais \\%\thanks{ Use footnote for providing further information about author (webpage, alternative address)---\emph{not} for acknowledging funding agencies.  Funding acknowledgements go at the end of the paper.} \\
University of Glasgow \\
11 Chapel Ln, Glasgow G11 6EW, UK\\
\small{\texttt{v.pais-malacalza.1@research.gla.ac.uk}} \\
\and
Malena Mendilaharzu \\
Dotphoton\\
Nordstrasse 3, 6300 Zug, Switzerland\\
\small{\texttt{malena.mendilaharzu@dotphoton.com}} \\
\and
Daniele Faccio \\
University of Glasgow\\
11 Chapel Ln, Glasgow G11 6EW, UK\\
\small{\texttt{daniele.faccio@glasgow.ac.uk}} \\
\and
Luis Oala \\
Dotphoton \& Brickroad \\
Nordstrasse 3, 6300 Zug, Switzerland\\
\small{\texttt{luis.oala@dotphoton.com}} \\
\and
Christoph Clausen \\
Dotphoton\\
Nordstrasse 3, 6300 Zug, Switzerland\\
\small{\texttt{christoph.clausen@dotphoton.com}} \\
\and
Bruno Sanguinetti \\
Dotphoton\\
Nordstrasse 3, 6300 Zug, Switzerland\\
\small{\texttt{bruno.sanguinetti@dotphoton.com}} \\
}

\begin{document}

\maketitle

\begin{abstract}
Real-world deployment of AI vision models is both fueled and limited by the data available for training and testing. Real datasets are sparse and uneven: long-tailed or unbalanced distributions hinder generalization, and the low number of samples in low density regions makes it hard to run evaluations. Synthetic data can fill these gaps, providing us with a way to sample the input space more continuously and improve data coverage for benchmarks. Focusing on the autonomous driving safety-critical case of pedestrian detection in the dark, we show how synthetic low-light samples can be used to better characterize the performance of a state-of-the-art object detection model as a function of the scene illumination. We use a synthetic RAW image augmentation technique to generate low-light samples that match the noise model of the camera sensor. Performance metrics on real and synthetic low-light data are similar, indicating that the AI model finds it hard to distinguish between them.
\end{abstract}

%\put(0,-390){\begin{minipage}{\textwidth}
\begin{figure*}[!t]
    \centering
    \includegraphics[width=\textwidth]{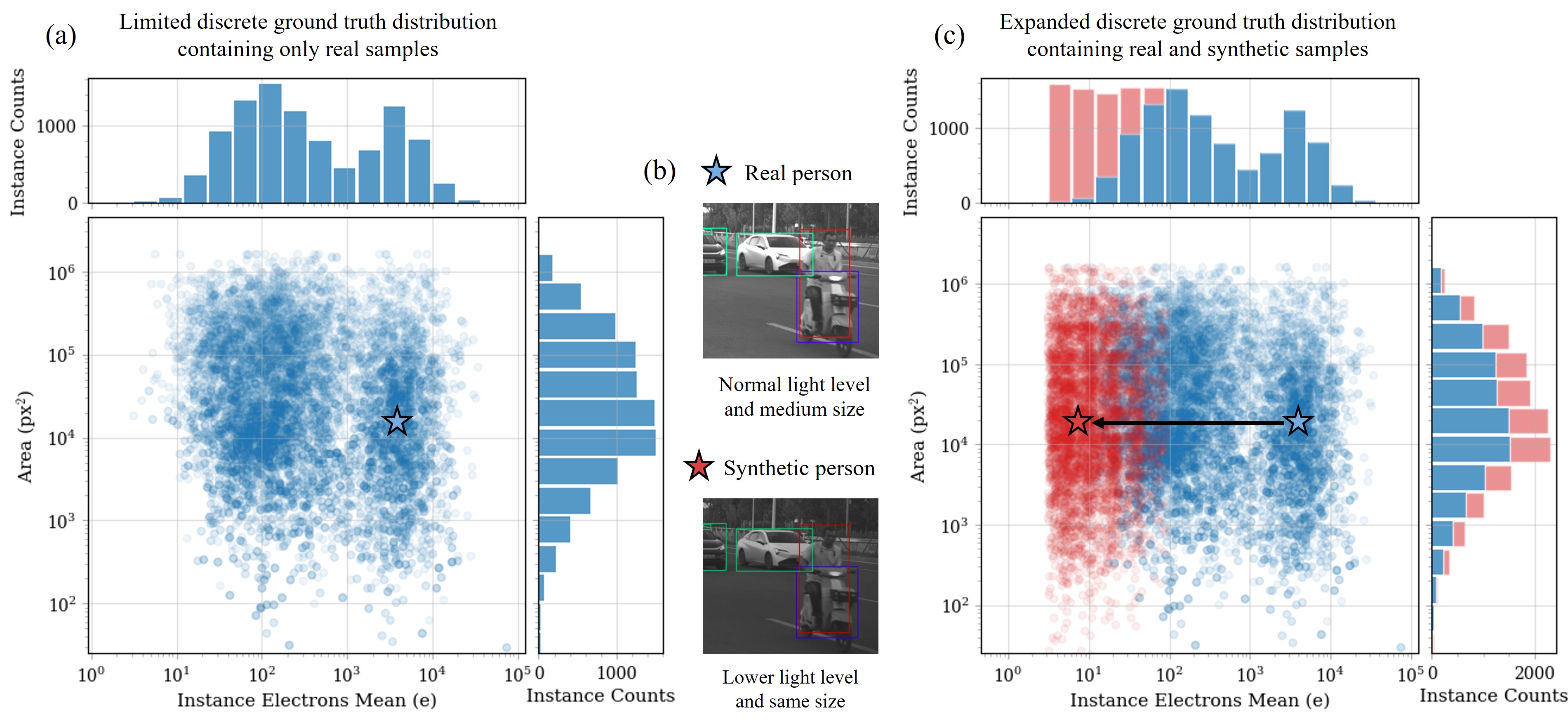}
    \vspace{-6mm}\caption{(a) AODRaw sliced test dataset distribution of person instances as a function of bounding box mean electrons and area. (b) Real normal-light sample from AODRaw and a synthetic example generated from it with a lower light level, but same area and noise model. (c) Guided by a complementary number of counts, we could sample target light levels from the resulting intensity PDF, synthesizing the data-points needed to create a more uniform overall distribution in the low light range, enabling a fairer evaluation.}
    \label{fig:real_gt_dist}
\end{figure*}
%\end{minipage}}

\vspace{-2mm}\section{Introduction}

Modern AI vision models have such a high performance that they can now act as the foundations of exciting autonomous agents such as self-driving cars. 
%Object detection algorithms trained or fine-tuned on real RAW images could improve the perception of such agents in dark scenes and other challenging conditions. 
However, vision AI models are both empowered and limited by the real data used to train them. Long-tail, unbalanced datasets and out-of-distribution (OOD) cases can become dangerous hazards while interacting with humans, such as failure to detect a pedestrian leading to accidents. We believe synthetic data can help evaluate and mitigate such hazards. 

Images express light measurements as a function of a 2D location, so vision models are bound to be affected by two main characteristics of objects: mean illumination and size. Low light levels can cause detection models to fail because of the small dynamic range inside the bounding box and the overall scene. It is also difficult to detect small objects, not only because objects further away are less illuminated, but also because their small size may converge to the spatial resolution itself. Having a low number of samples with low intensity and area (Fig \ref{fig:real_gt_dist}a) makes these data points a minority, and sometimes even OOD, further complicating their detection and evaluation.

Real data is discrete and limited, shaped by a data collection process that is expensive and naturally restricted by the scenes we can capture from the real world. In contrast, synthetic data allows a significantly higher degree of control. Owning the data generation pipeline, we can actively select samples to match target volumes and values on any continuous variable. For object detection models, we could in principle sample as many low-light scenes (Fig \ref{fig:real_gt_dist}b) as needed to balance the dataset (Fig. \ref{fig:real_gt_dist}c), allowing for broader data coverage when the model is trained and tested.

We apply this idea to better characterize the performance of a state-of-the-art object detection model. We first use real data to evaluate person detection metrics as a function of the scene illumination, while averaging over size to diminish its influence as a confounding factor. We then use real RAW images to generate synthetic low-light samples in a controlled environment where we repeat the performance analysis, gaining further insight that is now completely isolated from any size variation effects. We finally compare the performance of the model on real and synthetic data, discussing the validity of this approach.

\section{Related work}

%Real-world applications require safety guarantees not inherently offered by deep learning. 
Long-tailed distribution issues are severe in real-life applications such as autonomous driving because critical scenarios occur rarely and are dangerous to replicate \cite{2024ChenReview}. Synthetic generation through computer graphic simulations \cite{2017DosovitskiyCARLA,2023YangUniSim,2024ZhangChatScene} or generative AI \cite{2024LiGANAugmentation,2025BaiDcTDM,2025FengText2Traffic} raises questions on realism and fidelity because most of these methods do not fully consider the noise sources of CMOS sensors \cite{oala2023data}.
%Many of these techniques are either inconsistent with the 3D world, designed for sRGB images or other sensors such as LIDAR, or not realistic enough due to the lack of consideration for the noise sources of CMOS sensors.

%Synthetic data generation through computer graphics \TODO{cite} and image editing methods \TODO{cite} require critical scenarios to often be hand-crafted or prompted. Moreover, most generative approaches result on 2D images being inconsistent with the 3D world, and although physics-based methods address this, their computational cost is significantly high. More importantly, most of these synthetic generation methods are either designed for sRGB images or other sensors such as LIDAR, or do not account correctly for the noise sources of CMOS sensors.

Cameras possess an image signal processor (ISP) to generate sRGB images \cite{2019BrooksUnprocessing,2021DelbracioReview,2022CondeISP}. Unprocessed RAW images have a larger dynamic range and preserve more information along with the sensor's linearity and pixel noise independence. Training, pre-training or fine-tuning on RAW enables better object detection \cite{2025LiAODRaw}, instance \cite{2023ChenInstSegmentation} and semantic segmentation \cite{2024CuiRawAdapter} in low-light scenes. However, labeled RAW datasets are fewer and smaller than sRGB datasets (e.g. PASCAL RAW \cite{2015OmidZohoorPASCALRAW} and LOD \cite{2020HongLOD} have thousands of images, whereas ImageNet \cite{2009DengImageNet} has a million). 

%RGB pre-trained object detectors perform suboptimally on RAW images due to the domain gap \cite{2025BerdanReRAW}

%the quantity of RAW image datasets (i.e. 4259 images in PASCAL RAW [52] dataset and 2230 images in LOD [27] dataset) is incomparable to current RGB datasets (i.e. over 1M images in ImageNet [17] dataset and SAM [38] is trained with 11M images). \cite{2024CuiRawAdapter}

%Computational photography research 

To compensate for the lack of RAW data, many inverse ISP models aim at recovering RAW images from sRGB \cite{2019BrooksUnprocessing,2024Li,2025BerdanReRAW}. Some methods are designed to simultaneously tackle the domain gap between RAW and sRGB, adapting models trained on sRGB into RAW input domain \cite{2020ZamirCycleISP,2021XingInvISP,2024CuiRawAdapter}. 

An alternative solution is to carefully design data augmentation methods for RAW images that respect their noise and color filter array structure. RAW augmentations have been designed for cropping and clipping \cite{2019LiuBayerPreserving} and color jitter and blur \cite{2023YoshimuraRawgment}. We propose using Cui et al's \cite{2021CuiReduceIntensity} intensity reduction algorithm directly on RAW images as a light level RAW augmentation. Other Physics-based light level augmentations had been done before \cite{2018FeifanMBLLEN}, but only on sRGB, and without a full Poisson-Gaussian noise model.

%Research in computational photography and denoising takes the noise model into account. Most of the best performing inverse ISP models that recover RAW images from postprocessed sRGB images are nowadays learned end-to-end \TODO{\cite{2025BerdanReRAW}} instead of based on metadata \TODO{\cite{2019BrooksUnprocessing}}. Most inverse ISP methods are sensor-dependent and do not generalize well to other sensors. 

%\TODO{RAW augmentations \cite{2019LiuBayerPreserving,2023YoshimuraRawgment} + Summarize} %\vale{Mention that most inverse ISP methods are sensor-dependent and don't generalize well to other sensors. Talk about evaluation on unbalanced / long-tail datasets?}

\section{Methods}

\subsection{AODRaw dataset \& object detection models}

We use AODRaw \cite{2025LiAODRaw} as a study case: a state-of-the-art object detecion algorithm and dataset. 
%One of the best object detection models provided by the authors has a Cascade RCNN architecture \cite{2018CaiCascadeRCNN} with ConvNext \cite{2022LiuConvNext} 
%and Swin-T \cite{2021LiuSwin} backbones. 
Li et al report the best performance when first training on an ImageNet dataset \cite{2009DengImageNet} unprocessed into RAW images \cite{2019BrooksUnprocessing} and then finetuning on a collection of real RAW images made available inside the AODRaw dataset.

The AODRaw test subset contains 2260 $4024\times6048$ landscape images captured with a Sony A7M3 sensor. There are 4690 unique persons in these images, many of which are shown on traffic-related scenes and challenging conditions such as rain, fog and a low light setting. 

Li et al \cite{2025LiAODRaw} preprocess RAW images with two different strategies: downsampling and slicing. Although they obtain the best overall metrics training on downsampled images, the sliced pre-processing enables better detection of small objects in challenging conditions. Moreover, the downsampling operation heavily warps the noise properties of RAW images. We therefore limit ourselves to the 9650 person instances in sliced images.

For our experiments we build on the AODRaw repository and its MMDetetection interface \cite{2019ChenMMDetection}, along with its best-performing AODRaw model fine-tuned on sliced RAW: a Cascade RCNN model \cite{2018CaiCascadeRCNN} with ConvNext \cite{2022LiuConvNext} backbone. We calculate cumulative precision and recall to compute Average Precision (AP) \cite{2014LinCOCO}. We keep a constant score thresholds of 0.50 and we repeat the AP calculation for IOU thresholds between 0.50 and 0.95 with step 0.05. We then calculate Mean Average Precision (mAP) \cite{2014LinCOCO}.
%and Swin-T \cite{2021LiuSwin} backbones. 

%We use AODRaw \cite{2025LiAODRaw} as a case study. One of the best object detection models provided by the authors has a Cascade RCNN architecture \cite{2018CaiCascadeRCNN} with ConvNext \cite{2022LiuConvNext} 
%and Swin-T \cite{2021LiuSwin} backbones. 
%backbone. They report the best performance when first training on an ImageNet dataset \cite{2009DengImageNet} unprocessed into RAW images \cite{2019BrooksUnprocessing} and then finetuning on a collection of real RAW images made available inside the AODRaw dataseet.

%The test subset of the AODRaw dataset contains 2260 $4024\times6048$ landscape images captured with a Sony A7M3 sensor. There are 4690 unique pedestrian instances in the original images, many of which are shown on traffic-related scenes and challenging conditions such as rain, fog and a low light setting. 

%Li et al \cite{2025LiAODRaw} preprocess RAW images with two different strategies: downsampling and slicing. Although they obtain the best overall metrics training on downsampled images, the sliced preprocessing enables better detection of small objects in challenging conditions. Moreover, the downsampling operation heavily warps the noise properties of RAW images. We therefore limit ourselves to the 9650 person instances in sliced images.

\subsection{RAW electron maps and light level reduction}
\label{sec:augmentation}

RAW images contain direct measurements from the camera's sensor. In contrast to the sRGB assumption of Gaussian white noise, RAW images have signal-dependent noise. The Poisson-Gaussian model \cite{2008FoiPoissonGaussian} formalizes it in terms of Poisson shot noise -- i.e. signal variability due to the random time of arrival of photons -- and Gaussian readout noise caused by circuitry. A single heteroscedastic Gaussian distribution can be used as an approximation \cite{2014LiuNoise,2019BrooksUnprocessing}.

Simple digitization connects the RAW image array $y$ of pixel values in digital number (DNs) to the image array $x$ in physical units. We follow Hasinoff et al's \cite{2010HasinoffNoise} choice of units and we retrieve the number of electrons that flow through each pixel during the exposure time using Eq. \ref{eq:measurement}.

%RAW images contain the measurement information extracted directly from the camera. A simple digitization model (Eq. \ref{eq:measurement}) can be assumed to connect the image array $y$ of pixel values in digital number units (DNs) to the image array $x$ in physical units. We follow Hasinoff et al \cite{2010HasinoffNoise}, working with the number of electrons that flowed through each pixel during the exposure time.

%RAW images contain the measurement information extracted directly from the camera, stored in a digitized array $y$ expressed in digital number units (DNs). To retrieve the physical information, we assume a digitization model similar to Hasinoff et al's

%Following the choice of physical units from \cite{2010HasinoffNoise}, we express the image array $x$ as a map showing the number of electrons that flowed through each pixel during the exposure time. We assume we can retrieve $x$ from the stored array $y$ in digital number units (DNs) using Eq. \ref{eq:1}

%A simple digitization model can be assumed to connect the image array $y$ of pixel values in digital number units (DNs) to the image array $x$ in physical units. We can retrieve the number of electrons that flowed through each pixel during the exposure time using Eq. \ref{eq:measurement}

\vspace{-2mm}\begin{equation}
    y = gx + b + \epsilon
    \label{eq:measurement}
\end{equation}

Experimentation or manufacturers can provide us the overall gain $g$ in DN/e, the black level $b$ in DN and the average readout noise $\epsilon$ in DN needed to convert e into DNs, resulting on integers between 0 and the white level $2^d-1$ for usual bit depth $d=14$ for DNG files. ISO settings, found in the image metadata, control these parameters.

Working with RAW images makes it possible to design synthetic image pipelines grounded on Physics and the sensor's noise model. In this work we reduce the intensity of real images by retrieving their electron maps, thinning down their photon flux as done by Cui et al \cite{2024CuiRawAdapter}, and then reconverting to DNs. This results in a darker but otherwise identical synthetic image whose noise matches the real noise model. If needed, we can adjust the gain to preserve the general magnitude of the image in DNs before re-digitizing.

\section{Experiments and results}

\subsection{Classic evaluation limited by the amount of discrete real data available}
\label{sec:real_perf}

\begin{figure}[h]%[h]%[t!]
    \centering
    \includegraphics[width=0.46\textwidth]{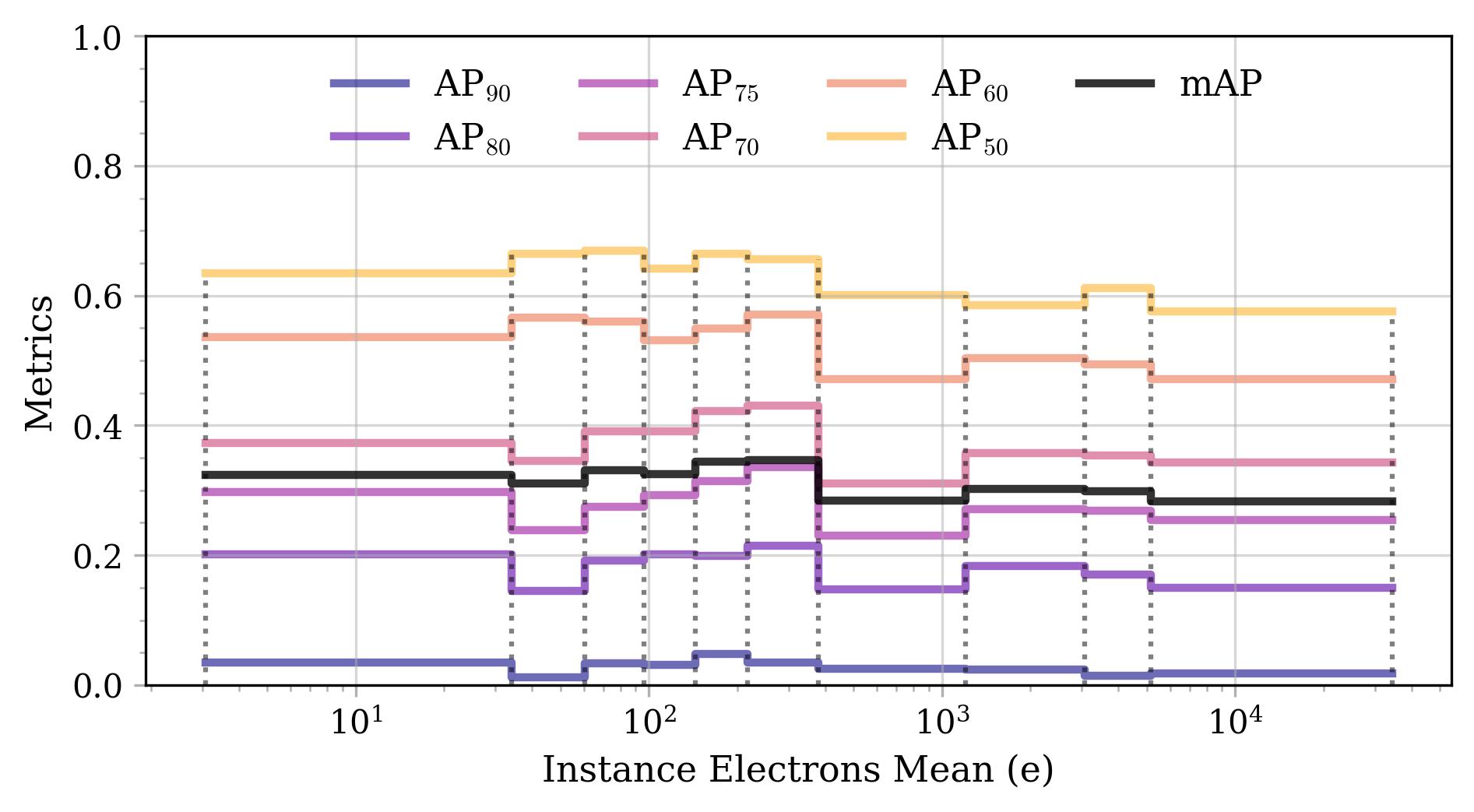}
    %\caption{Lorem ipsum, lorem ipsum,Lorem ipsum, lorem ipsum,Lorem ipsum}
    \vspace{-3mm}\caption{Person detection performance on real data as a function of instance illumination using approximately 500 TPs per interval. Numeric results can be found in the appendix in Table \ref{tab:real_intens}.}
    \label{fig:real_intens}
\end{figure}

Using all real test data, we studied person detection performance as a function of instance illumination (Fig \ref{fig:real_intens}). To estimate both precision and recall with similar confidence levels, we split the full range of light levels into 10 intervals that contained approximately the same number of true positives (TPs). Each interval resulted on a disjoint set of detections and ground-truth samples (GT) on which we evaluated metrics, independently characterizing each light level range while averaging over the entire range of area values.

%16_Test/Real_ConvNextSliced_Person_InstElectronsMean_TPCounts_20260517_163015

The low number of data points with very low light levels (Fig. \ref{fig:real_gt_dist}a) forced the first and last intervals to be large in logarithmic scale. Low density areas of the ground-truth distribution do not allow a finely-grained characterization. The highly-discrete nature of the data distribution sampled from reality limits the depth in which we can analyze the performance of vision models.

All metrics are approximately constant over the entire range of instance electron mean values available in the test subset (Fig. \ref{fig:real_intens}). This seems to indicate that AODRaw is as highly robust against changes in illumination as their authors claim \cite{2025LiAODRaw}. However, this is only true under the constraints of the real data collected inside the dataset. As seen in Fig. \ref{fig:real_gt_dist}a, less than 100 person instances have less than 100 electrons in average inside the bounding box. 

%Can we really estimate the performance at low light levels if we have such a low volume of data in this range?
% How would the model behave for even lower light levels? These are some of the questions that synthetic data could answer. 

\subsection{Finely-grained evaluation on synthetic data sampled from a latent continuous space}
\label{sec:synt_perf}

Synthetic data has the potential to enable a better evaluation of AI vision models, allowing for firmer control and higher discretization. We used the RAW light level augmentation described in Section \ref{sec:augmentation} to generate new samples and we repeated the performance evaluation as a function of light level exclusively on synthetic data. 

We chose a series of light level targets uniformly spaced in logarithm scale spanning the low light range of interest. We selected a group of 500 highly illuminated samples and we created one synthetic sample per real sample per target level. This methodology allows us to evaluate performance for different light levels on a fix number of samples with a very narrow spectrum of values of instance electron mean.

\begin{figure}[h]%[h]%[t!]
    \centering
    \includegraphics[width=0.46\textwidth]{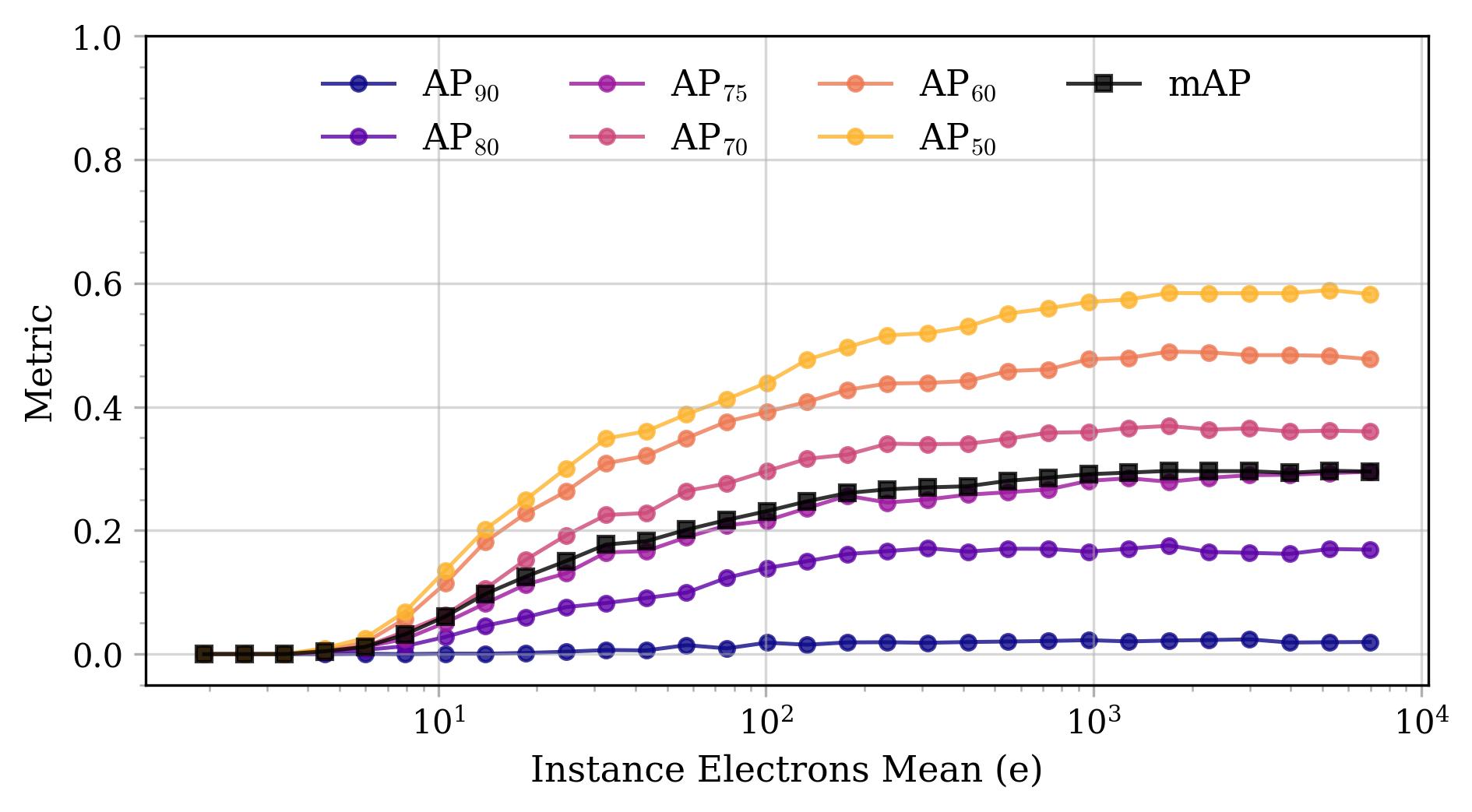}
    \vspace{-3mm}\caption{Person detection performance on synthetic data as a function of instance illumination using 500 GT samples per point. Numeric results can be found in the appendix in Table \ref{tab:synt_perf}.}
    \label{fig:synt_intens}
\end{figure}

Contrary to what real data evaluation suggested, our synthetic results indicate that the AODRaw detector does fail under very low scene illumination (Fig. \ref{fig:synt_intens}) whenever the gain or ISO settings are not adjusted accordingly. 

% 18_Full_Synt/Many_SlicedConvNext_Intens_Approx_500_30_person_20260421_105443

No detections were made with any of the IOU thresholds when the mean electrons count inside the bounding box is below 3.5 e. This caused AP metrics to become zero under 3.5 e. This threshold coincides with the minimum of the test dataset that makes any images with less than 3 e into OOD cases. The overall performance is deteriorated more slowly in the logarithmic scale, but mAP does decrease from 30\% to 10\% or less between 1000 and 10 e.

\subsection{Can the detector distinguish between real and synthetic data?}
\label{sec:undist}

%Our experiments show the benefits of using synthetic data for performance evaluation. 
Despite the theoretical assurance of the Physics-based noise model, the best way of testing the validity of our synthetic data is to experimentally determine whether real and synthetic data are undistinguishable to the model.

\begin{figure}[h]%[t!]
    \centering
    \begin{subfigure}{0.47\textwidth}
        \centering
        \includegraphics[width=0.93\textwidth]{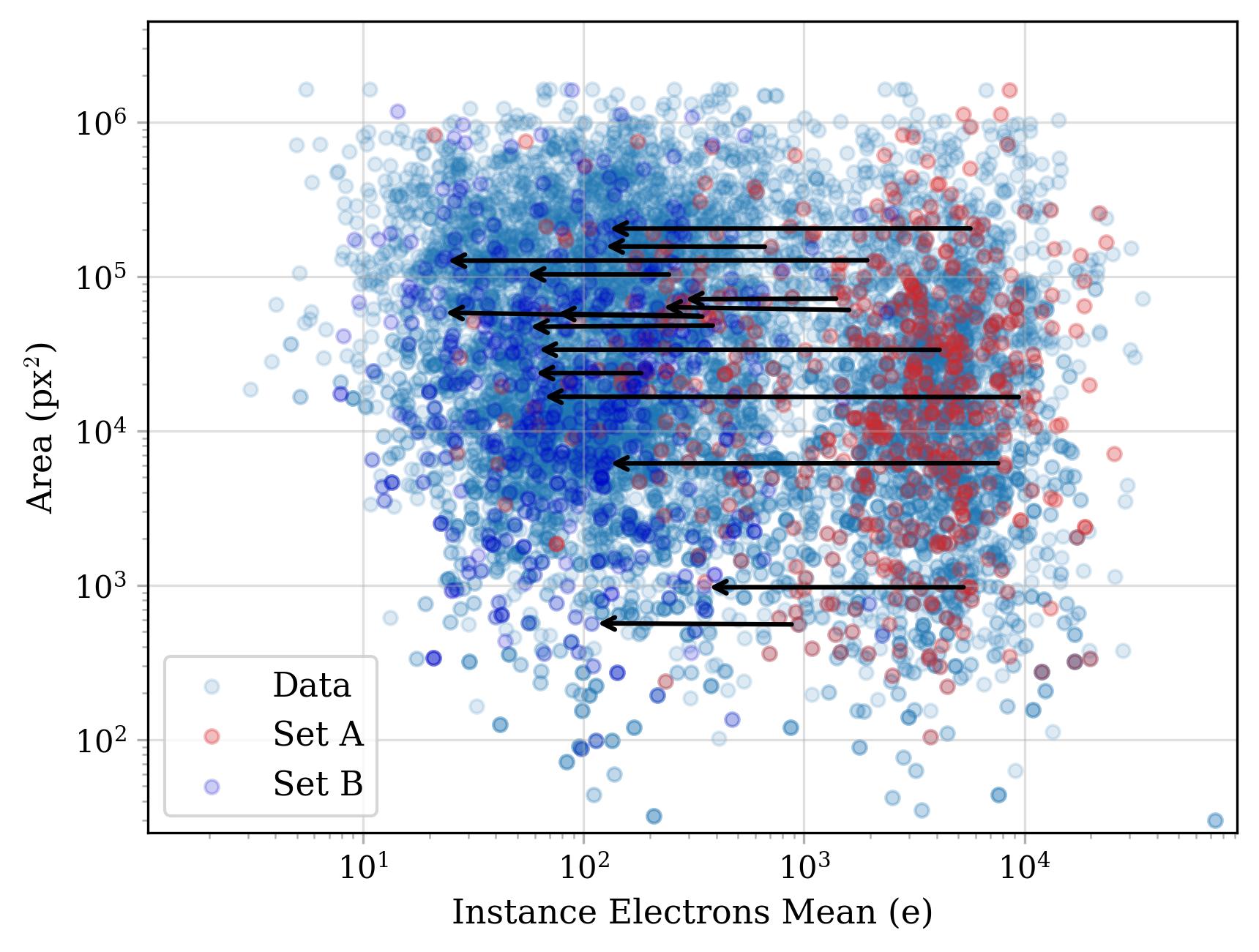}
    \end{subfigure}
    \begin{subfigure}{0.47\textwidth}
        \centering
        \includegraphics[width=\textwidth]{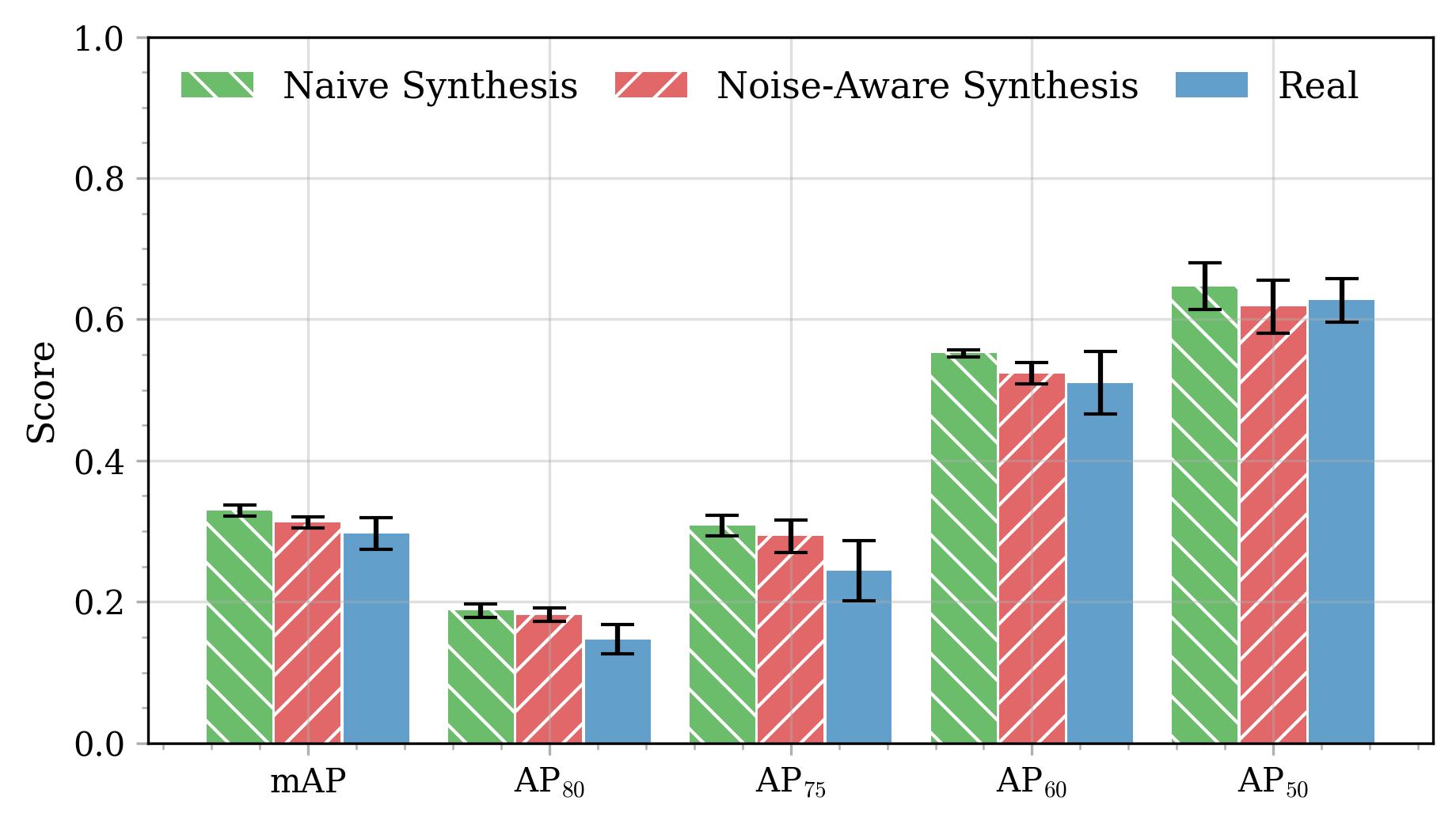}
        \label{subfig:synt_performance}
    \end{subfigure}
    \vspace{-6mm}\caption{Experiment to determine whether the detector can distinguish real and synthetic data. (a) Selected real person instances for high and low-light levels with arrows pointing from source to target. (b) Person detection performance on real and synthetic low-light data, comparing our noise-aware RAW augmentation with a simpler naive approach. Numeric results can be found in Table \ref{tab:ablation}}
    \label{fig:undist_performance} %Undist_AdjGain_SlicedConvNext_Inst_Intens_Approx_500_person_Multirun_3_20260412_133817
\end{figure}

We selected a set A of $N$ real person instances $a_1,...,a_N$ with high electron mean from the normal light category. We paired each $a_i$ to a real pedestrian $b_i$ with lower electron mean from the low light category (see Fig. \ref{fig:undist_performance}a). We did so minimizing the instance area difference between $a_i$ and $b_i$ to reduce the influence of area as a confounding factor.
%We forced instances to be unique inside of each set to avoid correlation between samples.

Our goal was to create a set C of $N$ synthetic pedestrians $c_i$ that had the low instance electron mean of a pedestrian $b_i$ but preserved most other characteristics from pedestrian $a_i$. For each $a_i$, we reduced the light level of the image so that its instance electron mean matched $b_i$'s. To avoid the influence of instance postprocessed mean in DN as a confounding factor, we adjusted the gain during re-digitization to match $b_i$'s ISO.

We evaluated performance on the real set B and compared it to the performance on the synthetic set C. We repeated the experiment 3 times with 6 disjoint sets A and B of $N=500$ data points (3000 points total). We report the average over these 3 independent runs with a single standard deviation error bar (see Fig. \ref{fig:undist_performance}b).% and Table \ref{tab:ablation}).

We repeated this procedure to compare our noise-aware RAW augmentation to a simpler naive approach. We generated a synthetic set $\tilde{\texttt{C}}$ using the same sets A and B, but we evenly reduced the intensity of each pixel on the electrons map without any noise adjustment.

%%17_Full_Undist/Ablation_20260517_203800

\vspace{-2mm}Similar mean values and overlapping error bars reveal that real and synthetic data are perceived in a similar way by the model (Fig. \ref{fig:undist_performance}b). Our noise-aware RAW augmentation performs consistently better than the naive approach, resulting on metrics closer to those of real data. Some person detection metrics (mAP, AP$_{75}$, AP$_{60}$) can only be considered statistically undistinguishable between real and synthetic data if the noise-aware method is used. 

High IOU thresholds (e.g. AP$_{80}$) result on non-overlapping error bars between real and synthetic metrics regardless of the method. This may indicate that the model can distinguish between real and synthetic data. However, it could also indicate that label quality in real dark scenes is not as good as in normal light. Our synthetic dark samples preserve the label quality from normal light and this might cause higher IOU between predicted and ground-truth bounding boxes.

%Our synthetic dark samples preserve the labelling quality from normal light samples, and this might cause more undetected instances and lower IOU, raising the number of false negatives and false positives.

% \begin{table}
% \centering
%     % \input{tables/ablation_four}
%     \input{tables/ablation_three}
%     \caption{Person detection metrics on real and synthetic data, comparing our noise-aware method and a naive synthesis to real data.}
%     \label{tab:ablation}
% \end{table}

\section{Conclusions}

In this publication we show how synthetic data can be used to improve the performance evaluation of vision models, effectively turning limited discrete variables into continuous controllable ones. 

We use a Physics-based light level RAW augmentation technique to generate synthetic low-light RAW images that follow the noise model from the camera sensor. Focusing on a safety-critical scenario for real-world applications such as autonomous driving, we design a controlled test environment to study person detection performance of a state-of-the-art object detection model. 

Available real data inside the test dataset is insufficient to provide proper detail to metric statistics, but a large enough number of synthetic samples can be created with target light levels in mind and greatly improve the characterization. Synthetic data is powerful, and using real RAW data as the source of synthetic data allows us to create samples that are undistinguishable from real samples according to most of the object detection model metrics.

%\subsubsection*{Author Contributions}
%If you'd like to, you may include  a section for author contributions as is done in many journals. This is optional and at the discretion of the authors.

\subsubsection*{Acknowledgments}

This work was supported by the UKRI EPSRC Centre for Doctoral Training in Applied Photonics [EP/S022821/1]. As part of this industry-academia collaboration program, the main author's PhD scholarship was co-funded by University of Glasgow and Dotphoton A.G.

{
    \small
    \bibliographystyle{ieeenat_fullname}
    \bibliography{main}
}

\ifshowappendix\appendix

\clearpage
\setcounter{page}{1}
\maketitlesupplementary

%\printphysics

\section{Real performance analysis}
\label{sec:real_appendix}

In this section, we provide methodology details for the real performance analysis from Section \ref{sec:real_perf}. We also include and discuss other results obtained by evaluating the model performance on real data.

%We studied several alternatives to bin the real data into non-overlapping intervals of instance mean electrons:

%For the TP binning strategy
We decided to study the scene illumination in terms of the mean number of electrons inside each instance's bounding box because it was a natural way of express the irradiance on each object or person. To define non-overlapping intervals that held the same number of TPs, we used IOU threshold 0.75 because AP$_{75}$ proved to be the closest AP metric to mAP in both real and synthetic data. 

We performed an analogous person performance analysis on real data as a function of other parameters. Among the most relevant was the instance area, whose results can be seen in Fig. \ref{fig:real_area}.

The performance drops considerably for persons whose instance area is below 14,500 px$^2$. This behavior was what originally led us to design experiments that account for the area, controlling this parameter whenever possible as it would be an impactful confounding factor.

\begin{figure}[h]%[h]%[t!]
    \centering
    \includegraphics[width=0.46\textwidth]{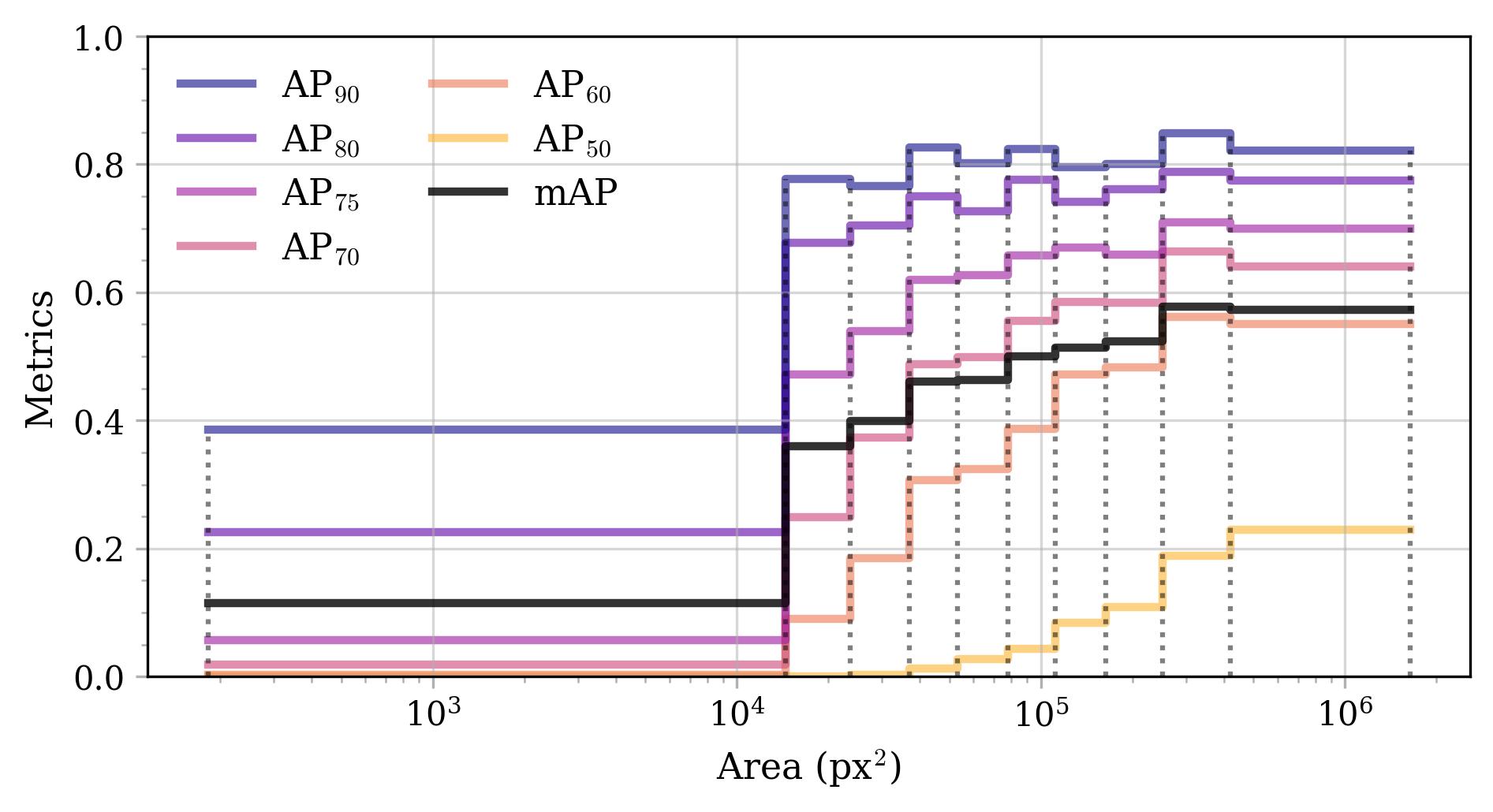}
    \vspace{-3mm}\caption{Person detection performance on real data as a function of instance bounding box area using approximately 400 TPs per interval. Numeric results can be found in Table \ref{tab:real_size}.}
    \label{fig:real_area}
\end{figure}
%16_Test/Real_ConvNextSliced_Person_Areas_TPCounts_20260517_205311

\vale{Could include metrics description}

\vale{Missing stuff for the thesis: other binning strategies + uncertainty estimation}

\section{Synthetic performance analysis}
\label{sec:synt_appendix}

We now provide methodology details for the synthetic performance analysis from Section \ref{sec:synt_perf}. In addition, we include and discuss results obtained with a gain-adjusted low-light RAW augmentation.

For data selection, we first filtered the sliced data according to class, keeping only person instances. We then applied an aspect ratio filter meant to avoid selecting badly sliced persons. For bounding box height $h$ and width $w$, we kept only those instances whose aspect ratio $h/w$ and inverse aspect ratio $w/h$ were both above the 90\% quantile of the full GT person distribution. Finally, we chose the $N=500$ instances with highest electron mean count.

Our simplest low light RAW augmentation adjusted the overall image mean in electrons to match a given target, making no changes to any ISO parameters before redigitizing the electrons map into a DN postprocessed output. This allowed us to not only investigate the performance under low light conditions, but to do so on non-ideal ISO conditions that are not originally present in the dataset.

To simulate the automatic ISO settings of the camera, we repeated the experiment from \ref{sec:synt_perf} but adjusting the gain $g$ to be inversely proportional to the light reduction factor $r$.

%\vspace{-1mm}
\begin{equation}
    I_f \approx r\, I_0 \; \Rightarrow \; g_f = \frac{g_0}{r}
    \label{eq:gain_adj}
\end{equation}

\vspace{2mm}We used the initial gain $g_0$ to obtain the electrons map, we reduced the intensity of each pixel by approximately a factor $r$ to make the final electrons mean $I_f$ approximately equal to $r\,I_0$, and we redigitized the electron map using an adjusted final gain $g_f$. Results are shown in Fig. \ref{fig:synt_intens_adj}.

\begin{figure}[b]%[h]%[t!]
    \centering
    \includegraphics[width=0.46\textwidth]{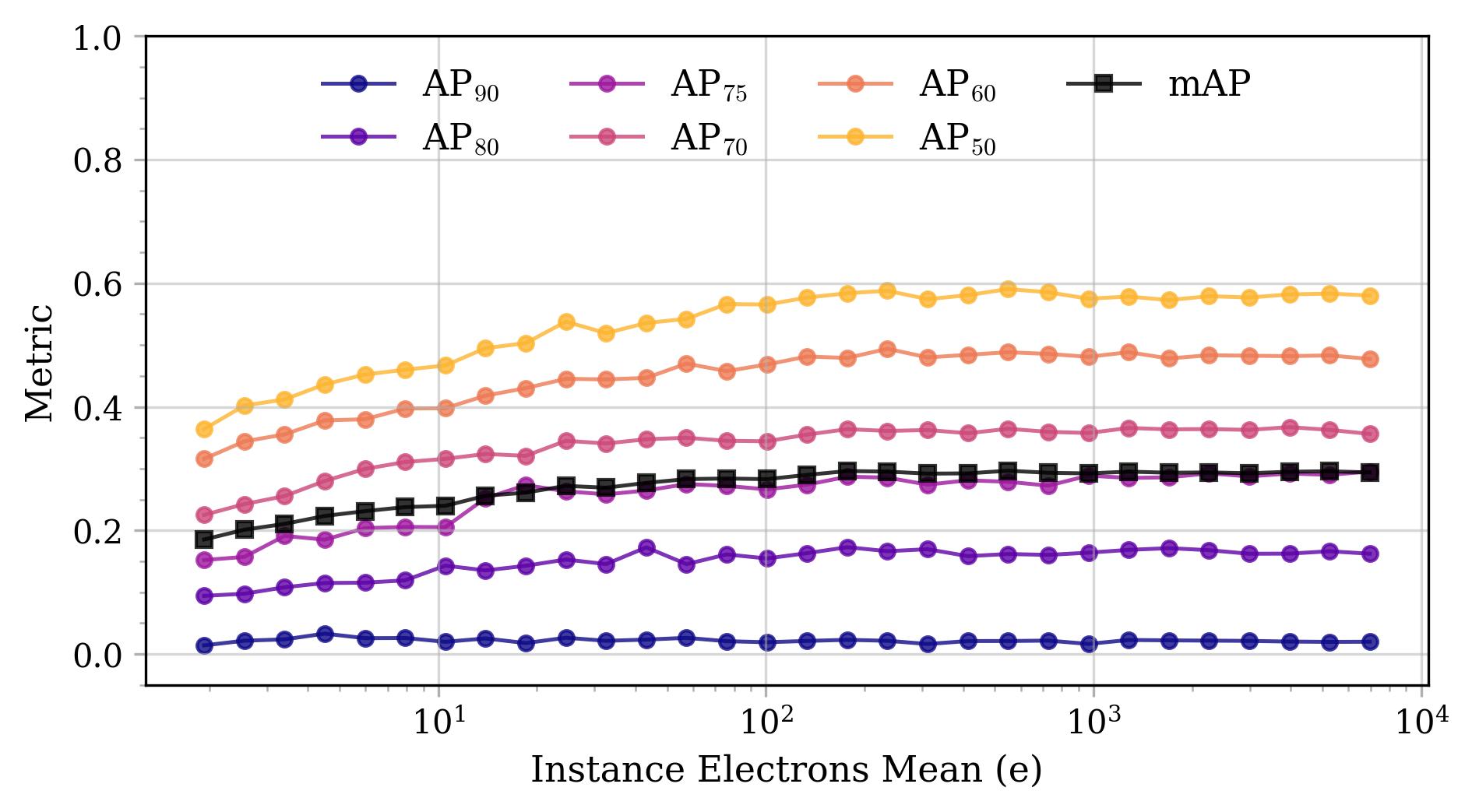}
    \vspace{-3mm}\caption{Person detection performance on synthetic data with gain adjustment as a function of instance illumination using 500 GT samples per point. Numeric results can be found in Table \ref{tab:synt_perf_adj}.}
    \label{fig:synt_intens_adj}
\end{figure}
% 18_Full_Synt/Many_AdjGain_SlicedConvNext_Intens_Approx_500_30_person_20260511_140721

The detection performance is still deteriorated by the light level reduction, falling by approximately 35\% between 235 and 2 e. However, the effect is not as marked as with no gain adjustment (see Fig. \ref{fig:synt_intens}). The AODRaw model AP performance metrics are never reduced to zero for decreasing illumination as long as the gain is adjusted to not have the postprocessed mean reduced proportionally to the unprocessed mean.

These results support two different conclusions. On the one hand, a lack of gain adjustment can partially explain the failure of the model, potentially indicating that low light is not such a problematic condition for object detection thanks to the work of Li et al \cite{2025LiAODRaw}. On the other hand, training on RAW data has not made the computer vision model ISO-invariant and this raises questions about the system's robustness. The gain is a simple multiplication factor applied to better store the information and it might not be desirable for it to have such an impact on downstream performance.

\vale{Missing stuff for the thesis: aspect ratio, overlap}

\section{Synthetic and real images comparison}
\label{sec:undist_appendix}

In this section, we provide details on the methodology used to compare the model performance on real and synthetic images in Section \ref{sec:undist}. We also discuss a more complete ablation study.

For data selection, we first applied the same class and aspect ratio filters described in \ref{sec:synt_appendix}. We split the resulting filtered dataset into low light and normal light subsets according to the sample tags provided by the AODRaw dataset. For each of these two subsets, we defined three random partitions with no intersection so we could independently run the experiment three times with no data repetition with a set A with $N=500$ normal light person instances and a set B with $M=750$ low-light person instances. We used a pool $1.5\times$ larger than needed to improve the matching quality. 

Pairing each element $a_i$ in A to an element $b_j$ in B was done with a Hungarian matching algorithm. We defined a cost matrix $L\in\mathbf{R}^{500\times750}$ such that $L_{ij} = s_{Ai} - s_{Bj}$ contained the difference in instance area $s$ between $a_i$ and $b_j$. We minimized the sum of cost matrix elements for chosen pairs with the linear sum assignment function from Scipy's Optimize module \cite{2020VitranenSciPy}, an implementation of a modified Jonker-Volgenant algorithm with no initialization \cite{2016CrouseAlgorithm}. This allowed us to define pairs with similar instance area, minimizing the influence of size as a confounding factor.

For a more complete ablation, we compared our noise-aware synthetic data pipeline with two other variants: (i) a naive non-adjusted synthesis that did not modify the gain nor any ISO setting parameter before redigitizing the electrons map into DNs, and (ii) a fully-adjusted noise-aware synthesis that modified not only the gain, but also the black level and the readout noise level to match those of the target image. Results are shown in Fig. \ref{fig:undist_perf_full}.

\begin{figure}[h]%[t!]
    \centering
    \includegraphics[width=0.48\textwidth]{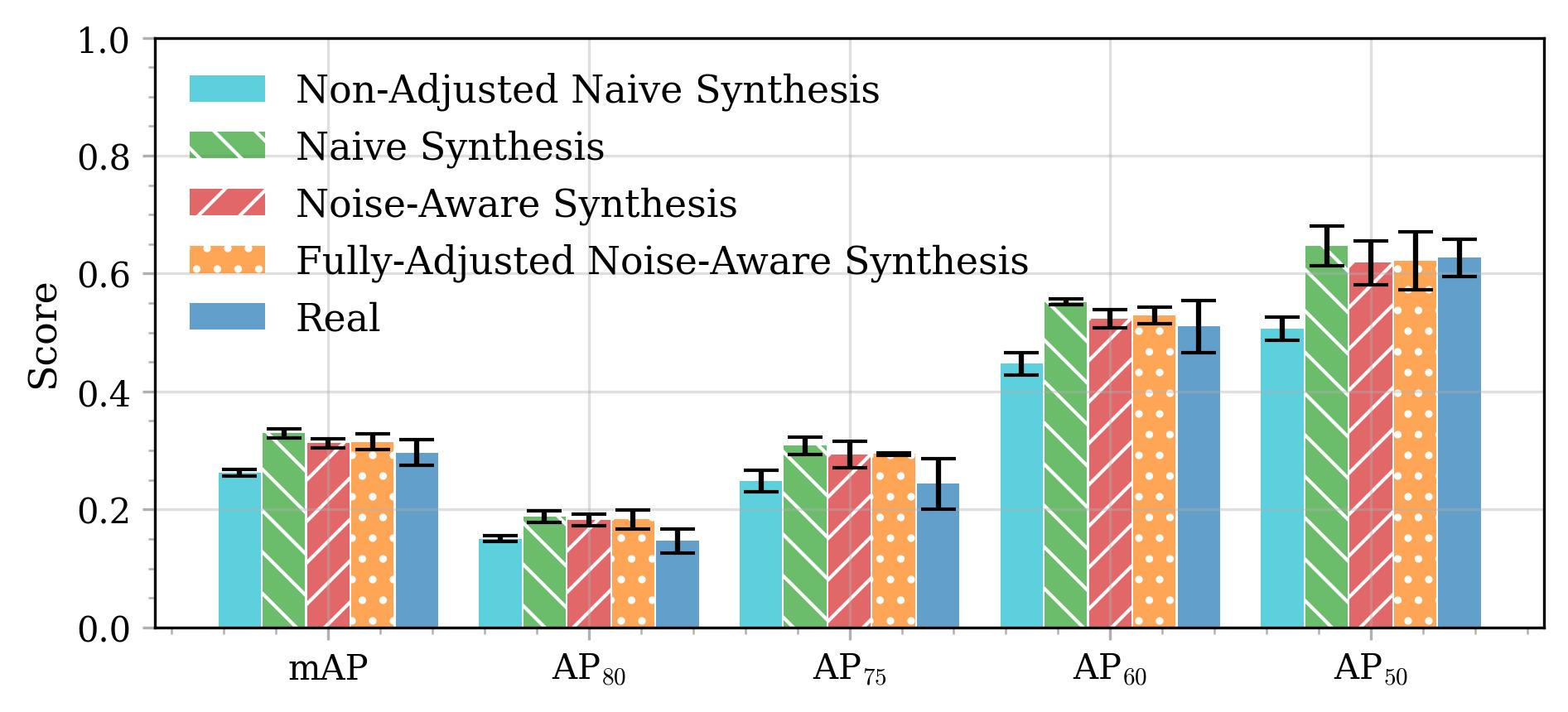}
    \vspace{-6mm}\caption{Person detection performance on real and synthetic low-light data, comparing our noise-aware RAW augmentation with three other synthetic data pipelines. Numeric results can be found in Table \ref{tab:ablation_full}}
    \label{fig:undist_perf_full} 
\end{figure}
%% 17_Full_Undist/Ablation_20260521_084331

Gain adjustment was proved fundamental for the model to similarly perceive real and synthetic data: the non-adjusted synthesis resulted on consistently lower AP metrics than the adjusted synthesis methods. The other two ISO setting parameters, the black level and the readout noise level, had very low impact on the way the model perceives the data. To decrease the synthetic pipeline overhead, it is reasonable to adjust only the gain.

\vale{Missing stuff for the thesis: Physics constraints}

\section{Numeric results for included figures}

To finalize this publication, we include numeric results for all of our figures in the following tables:

\vspace{2mm}\begin{itemize}
    \item Table \ref{tab:real_intens} contains results for the performance evaluation on real data as a function of person illumination covered in Section \ref{sec:real_perf} and Fig. \ref{fig:real_intens}.
    \item Table \ref{tab:synt_perf} lists results for the performance evaluation on synthetic data as a function of person illumination covered in Section \ref{sec:synt_perf} and Fig. \ref{fig:synt_intens}.
    \item Table \ref{tab:ablation} shows results for the experiment to determine whether the AODRaw model could distinguish real from synthetic data as explained in Section \ref{sec:undist} and Fig. \ref{fig:undist_performance}.
    \item Table \ref{tab:real_size} contains results for the performance evaluation on real data as a function of the instance area included in Section \ref{sec:real_appendix} and Fig. \ref{fig:real_area}.
    \item Table \ref{tab:synt_perf_adj} lists results for the performance evaluation on synthetic data with gain adjustment as a function of person illumination covered in Section \ref{sec:synt_appendix} and Fig. \ref{fig:synt_intens_adj}.
    \item Table \ref{tab:ablation_full} shows results for a more complete experiment on the AODRaw model performance on real and synthetic data created through different RAW augmentations, explained in Section \ref{sec:undist_appendix} and Fig. \ref{fig:undist_perf_full}.
\end{itemize}

\begin{table*}
\centering
    \footnotesize\begin{tabular}{ccccccccc}
\toprule
\multicolumn{2}{c}{\makecell[c]{Instance Mean}} & \multirow{2}{*}{mAP} & \multicolumn{6}{c}{AP} \\
\cmidrule(lr){1-2}\cmidrule(lr){4-9}
Min (e) & Max (e) & & AP$_{50}$ & AP$_{60}$ & AP$_{70}$ & AP$_{75}$ & AP$_{80}$ & AP$_{90}$ \\
\midrule
3 & 34 & 0.32 & 0.03 & 0.20 & 0.30 & 0.37 & 0.54 & 0.63 \\
34 & 60 & 0.31 & 0.01 & 0.14 & 0.24 & 0.35 & 0.57 & 0.66 \\
60 & 96 & 0.33 & 0.03 & 0.19 & 0.27 & 0.39 & 0.56 & 0.67 \\
96 & 145 & 0.33 & 0.03 & 0.20 & 0.29 & 0.39 & 0.53 & 0.64 \\
145 & 217 & 0.34 & 0.05 & 0.20 & 0.31 & 0.42 & 0.55 & 0.66 \\
217 & 378 & 0.35 & 0.03 & 0.21 & 0.34 & 0.43 & 0.57 & 0.66 \\
378 & 1203 & 0.28 & 0.03 & 0.15 & 0.23 & 0.31 & 0.47 & 0.60 \\
1203 & 3061 & 0.30 & 0.02 & 0.18 & 0.27 & 0.36 & 0.50 & 0.59 \\
3061 & 5152 & 0.30 & 0.01 & 0.17 & 0.27 & 0.35 & 0.49 & 0.61 \\
5152 & 34287 & 0.28 & 0.02 & 0.15 & 0.25 & 0.34 & 0.47 & 0.58 \\
\bottomrule
\end{tabular}
%16_Test/Real_ConvNextSliced_Person_InstElectronsMean_TPCounts_20260517_163015

    \caption{Person detection performance on real data as a function of instance illumination using approximately 500 TPs per interval.}
    \label{tab:real_intens}
\end{table*}

\begin{table*}
\centering
    \footnotesize\begin{tabular}{cccccccc}
\toprule
%\makecell[c]{Instance\\Mean (e)} & mAP & AP$_{50}$ & AP$_{60}$ & AP$_{70}$ & AP$_{75}$ & AP$_{80}$ & AP$_{90}$ \\
\multirow{2}{*}{\makecell[c]{Instance\\Mean (e)}} & \multirow{2}{*}{mAP} & \multicolumn{6}{c}{AP} \\
\cmidrule(lr){3-8}
 & & AP$_{50}$ & AP$_{60}$ & AP$_{70}$ & AP$_{75}$ & AP$_{80}$ & AP$_{90}$ \\
\midrule
%2 & 0.00 & 0.00 & 0.00 & 0.00 & 0.00 & 0.00 & 0.00 \\
%3 & 0.00 & 0.00 & 0.00 & 0.00 & 0.00 & 0.00 & 0.00 \\
3 & 0.00 & 0.00 & 0.00 & 0.00 & 0.00 & 0.00 & 0.00 \\
5 & 0.00 & 0.00 & 0.00 & 0.00 & 0.01 & 0.01 & 0.01 \\
6 & 0.01 & 0.00 & 0.01 & 0.01 & 0.01 & 0.02 & 0.03 \\
8 & 0.03 & 0.00 & 0.01 & 0.02 & 0.04 & 0.06 & 0.07 \\
11 & 0.06 & 0.00 & 0.03 & 0.05 & 0.06 & 0.11 & 0.14 \\
14 & 0.10 & 0.00 & 0.05 & 0.08 & 0.11 & 0.18 & 0.20 \\
18 & 0.13 & 0.00 & 0.06 & 0.11 & 0.15 & 0.23 & 0.25 \\
25 & 0.15 & 0.00 & 0.08 & 0.13 & 0.19 & 0.26 & 0.30 \\
33 & 0.18 & 0.01 & 0.08 & 0.16 & 0.23 & 0.31 & 0.35 \\
43 & 0.18 & 0.01 & 0.09 & 0.17 & 0.23 & 0.32 & 0.36 \\
57 & 0.20 & 0.01 & 0.10 & 0.19 & 0.26 & 0.35 & 0.39 \\
76 & 0.22 & 0.01 & 0.12 & 0.21 & 0.28 & 0.38 & 0.41 \\
101 & 0.23 & 0.02 & 0.14 & 0.22 & 0.30 & 0.39 & 0.44 \\
177 & 0.26 & 0.02 & 0.16 & 0.26 & 0.32 & 0.43 & 0.50 \\
235 & 0.27 & 0.02 & 0.17 & 0.25 & 0.34 & 0.44 & 0.52 \\
312 & 0.27 & 0.02 & 0.17 & 0.25 & 0.34 & 0.44 & 0.52 \\
413 & 0.27 & 0.02 & 0.17 & 0.26 & 0.34 & 0.44 & 0.53 \\
548 & 0.28 & 0.02 & 0.17 & 0.26 & 0.35 & 0.46 & 0.55 \\
727 & 0.29 & 0.02 & 0.17 & 0.27 & 0.36 & 0.46 & 0.56 \\
965 & 0.29 & 0.02 & 0.17 & 0.28 & 0.36 & 0.48 & 0.57 \\
1280 & 0.29 & 0.02 & 0.17 & 0.28 & 0.37 & 0.48 & 0.57 \\
1697 & 0.30 & 0.02 & 0.18 & 0.28 & 0.37 & 0.49 & 0.58 \\
2251 & 0.30 & 0.02 & 0.17 & 0.29 & 0.36 & 0.49 & 0.58 \\
2986 & 0.30 & 0.02 & 0.16 & 0.29 & 0.37 & 0.48 & 0.58 \\
3961 & 0.29 & 0.02 & 0.16 & 0.29 & 0.36 & 0.48 & 0.58 \\
5254 & 0.30 & 0.02 & 0.17 & 0.29 & 0.36 & 0.48 & 0.59 \\
6969 & 0.30 & 0.02 & 0.17 & 0.30 & 0.36 & 0.48 & 0.58 \\
\bottomrule
\end{tabular}
% 18_Full_Synt/Many_SlicedConvNext_Intens_Approx_500_30_person_20260421_105443
    \caption{Person detection metrics on synthetic data as a function of instance illumination using 500 GT samples per light level.}
    \label{tab:synt_perf}
\end{table*}

\begin{table*}
\centering
    \footnotesize\begin{tabular}{lccc}
\toprule
Metric & Naive Synthesis & Noise-Aware Synthesis & Real \\
\midrule
mAP & $0.33 \pm 0.01$ & $0.31 \pm 0.01$ & $0.30 \pm 0.02$ \\
AP$_{50}$ & $0.65 \pm 0.03$ & $0.62 \pm 0.04$ & $0.63 \pm 0.03$ \\
AP$_{55}$ & $0.61 \pm 0.02$ & $0.57 \pm 0.03$ & $0.58 \pm 0.03$ \\
AP$_{60}$ & $0.55 \pm 0.01$ & $0.52 \pm 0.02$ & $0.51 \pm 0.04$ \\
AP$_{65}$ & $0.48 \pm 0.01$ & $0.45 \pm 0.01$ & $0.42 \pm 0.03$ \\
AP$_{70}$ & $0.40 \pm 0.01$ & $0.38 \pm 0.01$ & $0.33 \pm 0.03$ \\
AP$_{75}$ & $0.31 \pm 0.01$ & $0.29 \pm 0.02$ & $0.24 \pm 0.04$ \\
AP$_{80}$ & $0.19 \pm 0.01$ & $0.18 \pm 0.01$ & $0.15 \pm 0.02$ \\
AP$_{85}$ & $0.08 \pm 0.02$ & $0.08 \pm 0.00$ & $0.08 \pm 0.02$ \\
AP$_{90}$ & $0.02 \pm 0.01$ & $0.02 \pm 0.00$ & $0.02 \pm 0.01$ \\
AP$_{95}$ & $0.00 \pm 0.00$ & $0.00 \pm 0.00$ & $0.00 \pm 0.00$ \\
\bottomrule
\end{tabular}
%17_Full_Undist/Ablation_20260517_203800
    \caption{Person detection metrics on real and synthetic data, comparing our noise-aware method and a naive synthesis to real data.}
    \label{tab:ablation}
\end{table*}

\begin{table*}
\centering
    \footnotesize\begin{tabular}{ccccccccc}
\toprule
\multicolumn{2}{c}{\makecell[c]{Instance Area}} & \multirow{2}{*}{mAP} & \multicolumn{6}{c}{AP} \\
\cmidrule(lr){1-2}\cmidrule(lr){4-9}
Min (px$^2$) & Max (px$^2$) & & AP$_{50}$ & AP$_{60}$ & AP$_{70}$ & AP$_{75}$ & AP$_{80}$ & AP$_{90}$ \\
\midrule
182 & 14490 & 0.11 & 0.39 & 0.23 & 0.06 & 0.02 & 0.00 & 0.00 \\
14490 & 23642 & 0.36 & 0.78 & 0.68 & 0.47 & 0.25 & 0.09 & 0.00 \\
23642 & 36864 & 0.40 & 0.77 & 0.70 & 0.54 & 0.37 & 0.18 & 0.00 \\
36864 & 53234 & 0.46 & 0.83 & 0.75 & 0.62 & 0.49 & 0.31 & 0.01 \\
53234 & 78003 & 0.46 & 0.80 & 0.73 & 0.63 & 0.50 & 0.32 & 0.03 \\
78003 & 111766 & 0.50 & 0.82 & 0.78 & 0.66 & 0.56 & 0.39 & 0.04 \\
111766 & 163796 & 0.51 & 0.80 & 0.74 & 0.67 & 0.58 & 0.47 & 0.08 \\
163796 & 250512 & 0.52 & 0.80 & 0.76 & 0.66 & 0.58 & 0.48 & 0.11 \\
250512 & 420246 & 0.58 & 0.85 & 0.79 & 0.71 & 0.66 & 0.56 & 0.19 \\
420246 & 1635841 & 0.57 & 0.82 & 0.77 & 0.70 & 0.64 & 0.55 & 0.23 \\
\bottomrule
\end{tabular}
%16_Test/Real_ConvNextSliced_Person_Areas_TPCounts_20260517_205311
    \caption{Person detection performance on real data as a function of instance area using approximately 400 TPs per interval.}
    \label{tab:real_size}
\end{table*}

\begin{table*}
\centering
    \footnotesize\begin{tabular}{cccccccc}
\toprule
%\makecell[c]{Instance\\Mean (e)} & mAP & AP$_{50}$ & AP$_{60}$ & AP$_{70}$ & AP$_{75}$ & AP$_{80}$ & AP$_{90}$ \\
\multirow{2}{*}{\makecell[c]{Instance\\Mean (e)}} & \multirow{2}{*}{mAP} & \multicolumn{6}{c}{AP} \\
\cmidrule(lr){3-8}
 & & AP$_{50}$ & AP$_{60}$ & AP$_{70}$ & AP$_{75}$ & AP$_{80}$ & AP$_{90}$ \\
\midrule
2 & 0.19 & 0.01 & 0.09 & 0.15 & 0.23 & 0.32 & 0.36 \\
3 & 0.20 & 0.02 & 0.10 & 0.16 & 0.24 & 0.34 & 0.40 \\
%3 & 0.21 & 0.02 & 0.11 & 0.19 & 0.26 & 0.36 & 0.41 \\
5 & 0.22 & 0.03 & 0.12 & 0.19 & 0.28 & 0.38 & 0.44 \\
6 & 0.23 & 0.03 & 0.12 & 0.20 & 0.30 & 0.38 & 0.45 \\
8 & 0.24 & 0.03 & 0.12 & 0.21 & 0.31 & 0.40 & 0.46 \\
11 & 0.24 & 0.02 & 0.14 & 0.21 & 0.32 & 0.40 & 0.47 \\
14 & 0.26 & 0.03 & 0.14 & 0.25 & 0.32 & 0.42 & 0.49 \\
18 & 0.26 & 0.02 & 0.14 & 0.27 & 0.32 & 0.43 & 0.50 \\
25 & 0.27 & 0.03 & 0.15 & 0.26 & 0.35 & 0.45 & 0.54 \\
33 & 0.27 & 0.02 & 0.15 & 0.26 & 0.34 & 0.44 & 0.52 \\
43 & 0.28 & 0.02 & 0.17 & 0.26 & 0.35 & 0.45 & 0.54 \\
57 & 0.28 & 0.03 & 0.15 & 0.28 & 0.35 & 0.47 & 0.54 \\
76 & 0.28 & 0.02 & 0.16 & 0.27 & 0.35 & 0.46 & 0.57 \\
101 & 0.28 & 0.02 & 0.15 & 0.27 & 0.34 & 0.47 & 0.57 \\
134 & 0.29 & 0.02 & 0.16 & 0.27 & 0.36 & 0.48 & 0.58 \\
177 & 0.30 & 0.02 & 0.17 & 0.29 & 0.36 & 0.48 & 0.58 \\
235 & 0.30 & 0.02 & 0.17 & 0.29 & 0.36 & 0.49 & 0.59 \\
312 & 0.29 & 0.02 & 0.17 & 0.27 & 0.36 & 0.48 & 0.57 \\
413 & 0.29 & 0.02 & 0.16 & 0.28 & 0.36 & 0.48 & 0.58 \\
548 & 0.30 & 0.02 & 0.16 & 0.28 & 0.36 & 0.49 & 0.59 \\
727 & 0.29 & 0.02 & 0.16 & 0.27 & 0.36 & 0.49 & 0.59 \\
965 & 0.29 & 0.02 & 0.16 & 0.29 & 0.36 & 0.48 & 0.57 \\
1280 & 0.30 & 0.02 & 0.17 & 0.28 & 0.37 & 0.49 & 0.58 \\
1697 & 0.29 & 0.02 & 0.17 & 0.29 & 0.36 & 0.48 & 0.57 \\
2251 & 0.29 & 0.02 & 0.17 & 0.29 & 0.36 & 0.48 & 0.58 \\
2986 & 0.29 & 0.02 & 0.16 & 0.29 & 0.36 & 0.48 & 0.58 \\
3961 & 0.30 & 0.02 & 0.16 & 0.29 & 0.37 & 0.48 & 0.58 \\
5254 & 0.30 & 0.02 & 0.17 & 0.29 & 0.36 & 0.48 & 0.58 \\
6969 & 0.29 & 0.02 & 0.16 & 0.30 & 0.36 & 0.48 & 0.58 \\
\bottomrule
\end{tabular}
    \caption{Person detection metrics on synthetic data with gain adjustment as a function of illumination using 500 GT samples per light level.}
    \label{tab:synt_perf_adj}
\end{table*}

\begin{table*}
\centering
    \footnotesize\begin{tabular}{lccccc}
\toprule
%\makecell[c]{Instance\\Mean (e)} & mAP & AP$_{50}$ & AP$_{60}$ & AP$_{70}$ & AP$_{75}$ & AP$_{80}$ & AP$_{90}$ \\
Metric & \makecell[c]{Non-Adjusted\\Naive\\Synthesis} & \makecell[c]{Naive \\Synthesis}& \makecell[c]{Noise-Aware\\Synthesis} & \makecell[c]{Fully-Adjusted\\Noise-Aware\\Synthesis} & Real \\
\midrule
mAP & $0.26 \pm 0.01$ & $0.33 \pm 0.01$ & $0.31 \pm 0.01$ & $0.31 \pm 0.01$ & $0.30 \pm 0.02$ \\
AP$_{50}$ & $0.51 \pm 0.02$ & $0.65 \pm 0.03$ & $0.62 \pm 0.04$ & $0.62 \pm 0.05$ & $0.63 \pm 0.03$ \\
%AP$_{55}$ & $0.48 \pm 0.01$ & $0.61 \pm 0.02$ & $0.57 \pm 0.03$ & $0.58 \pm 0.03$ & $0.58 \pm 0.03$ \\
AP$_{60}$ & $0.45 \pm 0.02$ & $0.55 \pm 0.01$ & $0.52 \pm 0.02$ & $0.53 \pm 0.01$ & $0.51 \pm 0.04$ \\
%AP$_{65}$ & $0.39 \pm 0.01$ & $0.48 \pm 0.01$ & $0.45 \pm 0.01$ & $0.45 \pm 0.02$ & $0.42 \pm 0.03$ \\
%AP$_{70}$ & $0.33 \pm 0.00$ & $0.40 \pm 0.01$ & $0.38 \pm 0.01$ & $0.38 \pm 0.01$ & $0.33 \pm 0.03$ \\
AP$_{75}$ & $0.25 \pm 0.02$ & $0.31 \pm 0.01$ & $0.29 \pm 0.02$ & $0.29 \pm 0.00$ & $0.24 \pm 0.04$ \\
AP$_{80}$ & $0.15 \pm 0.01$ & $0.19 \pm 0.01$ & $0.18 \pm 0.01$ & $0.18 \pm 0.02$ & $0.15 \pm 0.02$ \\
AP$_{85}$ & $0.06 \pm 0.02$ & $0.08 \pm 0.02$ & $0.08 \pm 0.00$ & $0.08 \pm 0.01$ & $0.08 \pm 0.02$ \\
%AP$_{90}$ & $0.01 \pm 0.00$ & $0.02 \pm 0.01$ & $0.02 \pm 0.00$ & $0.02 \pm 0.00$ & $0.02 \pm 0.01$ \\
%AP$_{95}$ & $0.00 \pm 0.00$ & $0.00 \pm 0.00$ & $0.00 \pm 0.00$ & $0.00 \pm 0.00$ & $0.00 \pm 0.00$ \\
\bottomrule
\end{tabular}
% 17_Full_Undist/Ablation_20260521_084331
    \caption{Person detection metrics on real and synthetic data, comparing our noise-aware method and three other synthetic data pipelines.}
    \label{tab:ablation_full}
\end{table*}

\section{Licences}

AODRaw code and pre-trained models \cite{2025LiAODRaw} have been published under the Creative Commons BY-NC-SA 4.0 license. The MMDetection code suite \cite{2019ChenMMDetection} can be used under Apache 2.0 license. Scipy \cite{2020VitranenSciPy} is an open-source package under the BSD-3-Clause license.\fi

% WARNING: do not forget to delete the supplementary pages from your submission 
% \input{sec/X_suppl}

\end{document}